\documentclass[runningheads]{llncs}
\usepackage{graphicx}
\usepackage{amsmath,amsfonts,amssymb}
\usepackage{color}
\usepackage[width=122mm,left=12mm,paperwidth=146mm,height=193mm,top=12mm,paperheight=217mm]{geometry}
\usepackage[utf8x]{inputenc}
\usepackage[lined,boxed,commentsnumbered]{algorithm2e}
\usepackage{booktabs}
\usepackage{subfigure}
\usepackage{soul}

\font\dsrom=dsrom10 scaled 1200
\newcommand{\indicator}[1]{\textrm{\dsrom{1}}_{#1}}

\begin{document}

\pagestyle{headings}
\mainmatter

\title{Weakly Supervised Action Labeling in Videos Under Ordering Constraints}


\author{
Piotr~Bojanowski\textsuperscript{1}\thanks{WILLOW project-team, DI/ENS, ENS/INRIA/CNRS UMR 8548, Paris, France.}
\;\;\;
R\'emi~Lajugie\textsuperscript{1}\thanks{SIERRA project-team, DI/ENS, ENS/INRIA/CNRS UMR 8548, Paris, France.}
\;\;\;
Francis~Bach\textsuperscript{1}$^{\star\star}$
\;\;\;
Ivan~Laptev\textsuperscript{1}$^{\star}$
\;\;\; \\
Jean~Ponce\textsuperscript{2}$^{\star}$
\;\;\;
Cordelia~Schmid\textsuperscript{1}\thanks{LEAR team, INRIA Grenoble Rh\^one-Alpes, France.}
\;\;\;
Josef~Sivic\textsuperscript{1}$^\star$
\
}
\institute{
\textsuperscript{1}INRIA \quad \textsuperscript{2}\'{E}cole Normale Sup\'erieure
}

\authorrunning{Piotr Bojanowski et al.}

\maketitle

\begin{abstract}
We are given a set of video clips, each one annotated with an {\em ordered} list of actions, such as ``walk'' then ``sit" then ``answer phone'' extracted from, for example, the associated text script.
We seek to temporally localize the individual actions in each clip as well as to learn a discriminative classifier for each action.
We formulate the problem as a weakly supervised temporal assignment with ordering constraints. Each video clip is divided into small time intervals and each time interval of each video clip is assigned one action label, while respecting the order in which the action labels appear in the given annotations.
We show that the action label assignment can be determined together with learning a classifier for each action in a discriminative manner.
We evaluate the proposed model on a new and challenging dataset of 937 video clips with a total of 787720 frames containing sequences of 16 different actions from 69 Hollywood movies.
\end{abstract}

\section{Introduction}

\begin{figure}[t]
    \centering
    \includegraphics[width=0.95\linewidth]{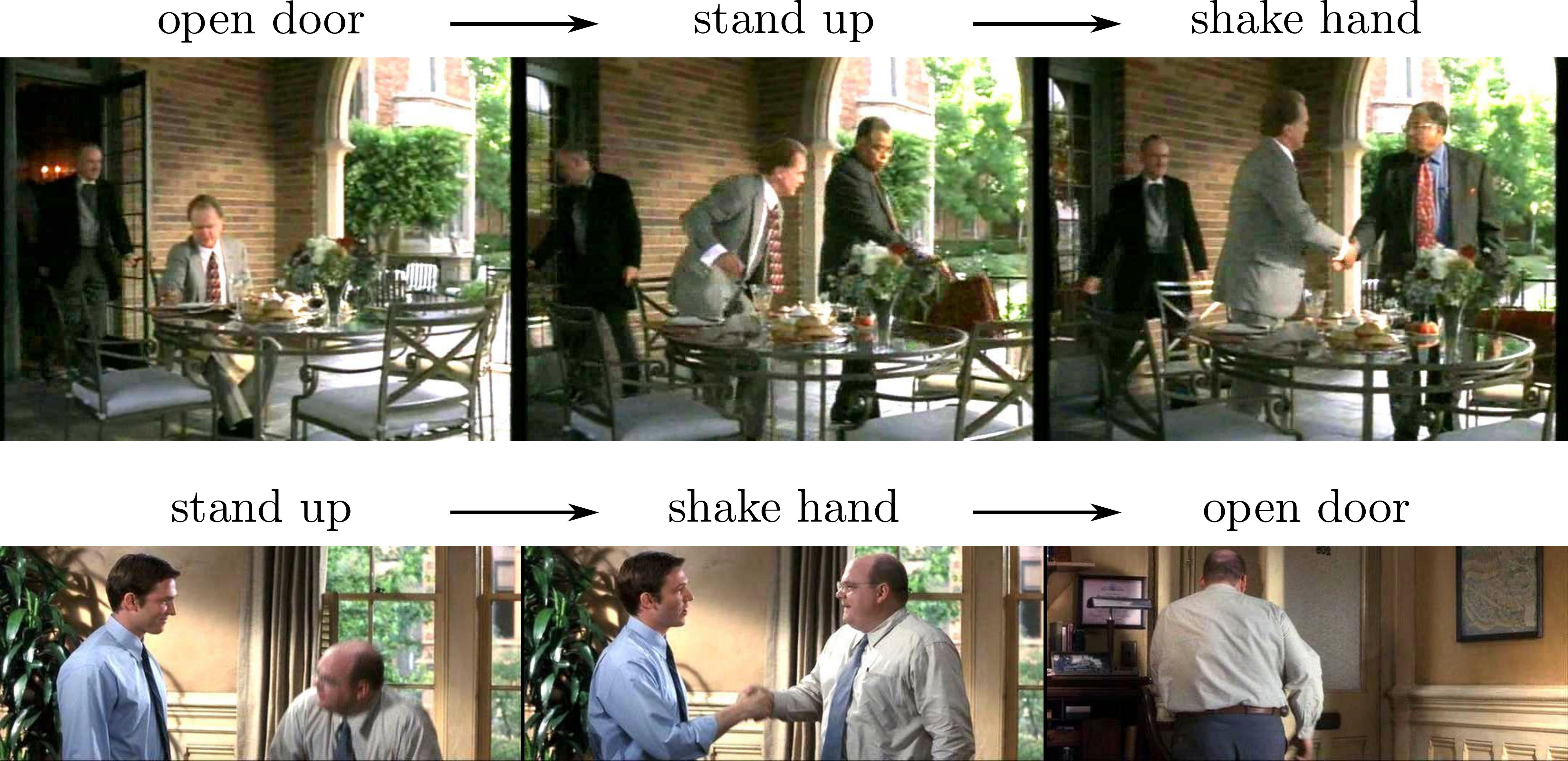}
    \caption{
        Examples of video clips with associated actions sequence annotations such as provided in our dataset.
        Both examples contain the same set of actions but occurring in a different order.
        In this work we use the type and order of events as a supervisory signal to learn a classifier of each action and temporally localize each action in the video. 
    }
    \label{fig:intro}
\end{figure}

Significant progress towards action recognition in realistic video settings has been achieved in the past few years~\cite{Laptev08a,Liu11,Niebles10a,Sadanand12a,Wang11a}. 
However action recognition is often cast as a classification or detection problem using fully annotated data, where the temporal boundaries of individual actions, e.g.\ in the form of pre-segmented video clips, are given during training.
The goal of this paper is to exploit the supervisory power of the temporal ordering of actions in a video stream, as illustrated in figure~\ref{fig:intro}. 

Gathering fully annotated videos with accurately time-stamped action labels is quite time consuming in practice. 
This limits the utility of fully supervised machine learning techniques on large-scale data.
Using data redundancy, weakly and semi-supervised methods are a promising alternative in this case. 
On the other hand, it is easy to gather videos with some level of textual annotation but poor temporal localization, from movie scripts for example.
This type of weak supervisory signal has been used before in classification~\cite{Laptev08a} and temporal localization~\cite{Duchenne09a} tasks.  
However, the crucial  information on the ordering of actions has, to the best of our knowledge, been ignored so far in the weakly supervised setting.
Following recent work on discriminative clustering \cite{Bach07diffrac,Xu2004maximum}, image \cite{Joulin10discriminative} and video \cite{Bojanowski13finding} cosegmentation, we propose to exploit this information in a discriminative framework where both the action model and the optimal assignments under temporal constraints are learned together.

\subsection{Related Work}\label{sec:relatedWork}

{\bf The temporal ordering of actions}, e.g.\ in the form of Markov models or action grammars, have been used to constrain action prediction in videos~\cite{Hongeng03large,Ivanov00recognition,Kwak11scenario,Laxton07leveraging,Ryoo06recognition,Vu03automatic}.
These kinds of spatial and temporal constraints have been also used in the context of group activity recognition~\cite{Amer13monte,Khamis12combining}.
Similar to us, these papers exploit  the temporal structure of videos, but focus on  inferring action sequences from noisy but pre-defined action detectors,
often in constrained surveillance and laboratory settings with a limited number of actions and static cameras.
In contrast, in this work we explore the temporal structure of actions for learning action classifiers in a weakly supervised set-up
and show results on challenging videos from feature length movies.
 
Related is also work on recognition of {\bf composite activities}~\cite{rohrbach12script}, where atomic action models (``cut'', ``open'') are learned given full supervision on a cooking video dataset.
Composite activity models (``prepare pizza'') are learned on top of the atomic actions, using the prediction scores for the atomic actions as features.
Annotations are, however, used without taking into account the ordering of actions.

{\bf Temporal models} for recognition {\bf of individual actions} have been explored in e.g.~\cite{Laptev08a,Niebles10a,Tang12}.
Implicit models in the form of temporal pyramids have been used with bag-of-features representations~\cite{Laptev08a}. 
Others have used more explicit temporal models in the form of, e.g. latent action parts~\cite{Niebles10a} or hidden Markov models~\cite{Tang12}. 
Contrary to these methods, we do not use an a priori model of the temporal structure of individual actions, but instead exploit the given ordering
constraints between actions to learn better individual actions models.

{\bf Weak supervision for learning actions} has been explored in~\cite{Bojanowski13finding,Duchenne09a,Laptev08a}. 
These methods use uncertain temporal annotations of actions provided by movie scripts.
Contrary to these works our method learns multiple actions simultaneously and incorporates temporal ordering constraints on action labels obtained, e.g. from the movie scripts.

{\bf Dynamic time warping algorithms (DTW)} can be used to match temporal sequences, and are extensively used in speech recognition, e.g.~\cite{Gold11speech,Rabiner93fundamentals}.
In computer vision, the temporal order of events has been exploited in~\cite{hoai11joint}, where a DTW-like algorithm is used at test time to improve the performance of non-maximum suppression on the output of pre-trained action detectors.

{\bf Discriminative clustering} is an unsupervised method that partitions data by minimizing a discriminative objective, optimizing over both classifiers and labels~\cite{Bach07diffrac,Xu2004maximum}.
Convex formulations of discriminative clustering have been explored in~\cite{Bach07diffrac,Guo2007convex}.
In computer vision these methods have been successfully applied to co-segmentation~\cite{Joulin12a}.
The approach presented in this paper is inspired by this framework, but adds to it the use of ordering constraints. 

In this work, we make use of the {\bf Frank-Wolfe algorithm} (a.k.a conditional gradient) to minimize our cost function.
The Frank-Wolfe algorithm~\cite{Frank1956,Jaggi2013} is a classical convex optimization procedure that permits optimizing a continuously differentiable convex function over a convex compact domain only by optimizing linear functions over the domain.
In particular, it does not require any projection steps. 
It has recently received increased attention in the context of large-scale optimization \cite{Harchaoui2012conditional,Jaggi2013}.

\subsection{Problem Statement and Contributions}

\begin{figure}[t]
    \centering
    \includegraphics[width=\linewidth]{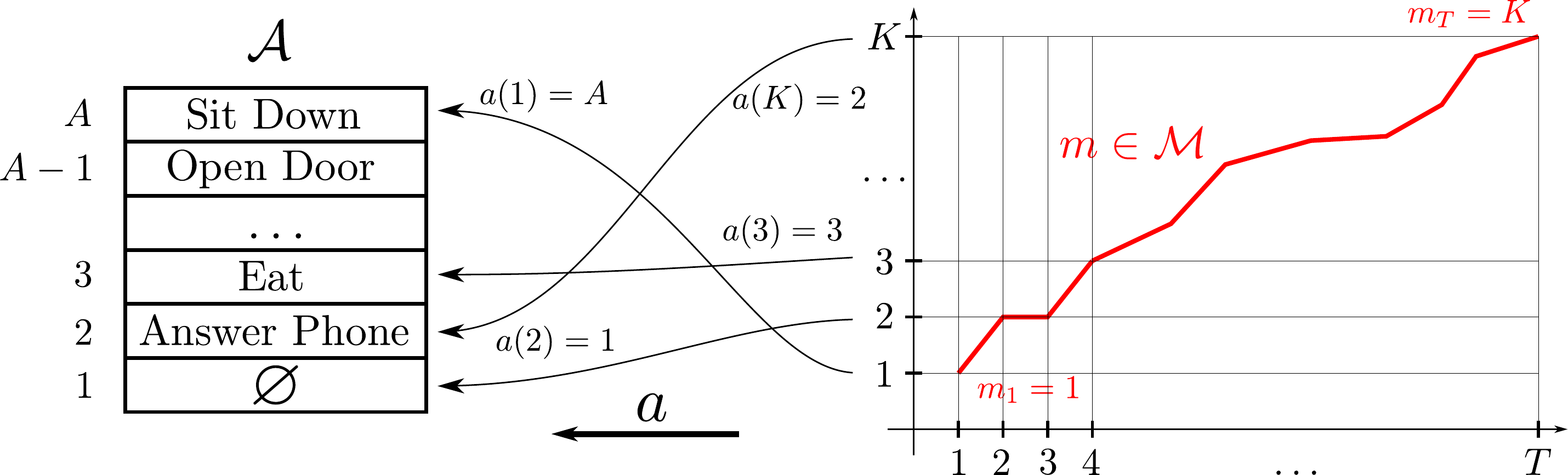}
    \caption{   
	Right: The goal is to find assignment of video intervals $1$ to $T$ (x-axis) to the ordered list of action annotations $a(1)$ to $a(K)$ indexed by integer $k$ from $1$ to $K$ (y-axis).
        Left: The ordered annotation index $k$ is mapped, through mapping $a$ to action labels from the set $\mathcal{A}$, in that example $a(3)=$``Eat''.
        To preserve the given ordering of annotations we only consider assignments $\mathcal{M}$ that are non-decreasing. One such assignment $m$ is shown in red.
    }
    \label{fig:warping}
\end{figure}

The temporal assignment problem addressed in the rest of this paper and illustrated by Fig.~\ref{fig:intro} can be stated as follows: We are given a set of $N$ video clips (or clips for short in what follows). 
A clip is defined as a contiguous video segment consisting of $F$ frames, and may correspond, for example, to a scene (as defined in a movie script) or a collection of subsequent shots.  
Each clip is divided into $T$ small \emph{time intervals} (chunks of videos consisting of $F/T=10$ frames in our case), and annotated by an ordered list of $K$ elements taken from some action set ${\cal A}$ of size $A = |{\cal A}|$ (that may consist of labels such as ``open door'', ``stand up'', ``answer phone'', etc., as in Fig.~1 for example). 
Note that clips are not of the same length but for the sake of simplicity, we assume they are.
We address the problem of assigning to each time interval of each clip one action in ${\cal A}$, respecting the order in which the actions appear in the original annotation list (Fig.~\ref{fig:warping}).

\paragraph{Contributions.}
We make the following contributions: 
\textbf{(i)} we propose a discriminative clustering model (section~\ref{sec:formulation}) that handles weak supervision in the form of temporal ordering constraints and recovers a classifier for each action together with the temporal localization of each action in each video clip;   
\textbf{(ii)} we design a convex relaxation of the proposed  model and  show it can be efficiently solved using the conditional gradient (Frank-Wolfe) algorithm (section~\ref{sec:optim}); 
and finally \textbf{(iii)} we demonstrate improved performance of our model on a new action dataset for the tasks of temporal localization (section~\ref{sec:results}) and action classification (section~\ref{sec:expClassif}). 
All the data and code are publicly available at \texttt{http://www.di.ens.fr/willow/research/ordering}.

\section{Discriminative Clustering with Ordering Constraints}\label{sec:formulation}

In this section we describe the proposed discriminative clustering model that incorporates label ordering constraints.
The input is a set of video clips, each annotated with an ordered list of action labels specifying the sequence of actions present in the clip.
The output is the temporal assignment of actions to individual time intervals in each clip respecting the ordering constraint provided by the annotations
together with a learnt classifier for each action, common for all clips.  In the following, we first formulate the temporal assignment of actions to individual frames as  discriminative clustering (section~\ref{sec:problem}), then introduce a parametrization of temporal assignments using indicator variables (section~\ref{sec:indicators}), 
and finally we describe the choice of a loss function for the discriminative clustering that leads to a convex cost (section~\ref{sec:cost}).  


\subsection{Problem Formulation}
\label{sec:problem}
Let us now formalize the temporal assignment problem.  
We denote by $x_n(t)$ in $\mathbb{R}^d$ some local descriptor of video clip number $n$ during time interval number $t$. 
For every $k$ in $\{1,\ldots,K\}$, we also define $a_n(k)$ as the element of ${\cal A}$ corresponding to annotation number $k$ (Fig.~\ref{fig:warping}). 
Note that the set of actions ${\cal A}$ itself is not ordered: even if we represent ${\cal A}$ by a table for convenience, the elements of this table are action labels and have no natural order. 
The annotations, on the other hand, are ordered, for example according to where they occur in a movie script, and are represented by some integer between $1$ and $K$. 
Thus $a_n$ maps (ordered) annotation indices onto (unordered) actions, and depends of course on the video clip under annotation.
Parts of any video clip may belong to the background.
To account for this fact, a dummy label $\varnothing$ is inserted in the annotation list between every consecutive pair of actual labels.

Let us denote by \(\mathcal{M}\) the set of \emph{admissible assignments} on \(\{1,\dots,T\}\), that is, the set of sequences \(m = (m_1,\ldots,m_T)\) with elements in \(\{1, \dots, K\}\) such that \(m_1=1\), \(m_T=K\), and \(m_{t+1}=m_t\) or \(m_{t+1}= m_t+1\) for all \(t\) in \(\{1, \dots,T-1\}\)
Such an assignment is illustrated in Fig.~\ref{fig:warping}. 

Let us also denote by \(\mathcal{F}\) the space of classifiers of interest, by \(\Omega : \mathcal{F} \to \mathbb{R}\) some regularizer on this space and by \( \ell : \mathcal{A} \times \mathbb{R}^A \to \mathbb{R}_+\) an appropriate loss function.
For a given clip \(n\) and a fixed classifier \(f\), the problem of assigning the clip intervals to the annotation sequence can be written as the minimization of the cost function:  
\begin{equation}
    E(m,f,n) = \frac{1}{T} \sum_{t=1}^{T} \ell \left ( a_n(m_t), f(x_n (t)) \right ) 
    \label{eq:E}
\end{equation}
with respect to assignment \(m\) in \(\mathcal{M}\).
The regularizer \(\Omega\) prevents overfitting and we therefore define a scalar parameter \(\lambda\) to control this effect.
Jointly learning the classifiers and solving the assignment problem corresponds to the following optimization problem: 

\begin{equation}
    \min_{f \in \mathcal{F}} \ \left [ \sum_{n=1}^N \ \min_{m \in \mathcal{M}} \ E(m,f,n) \right ] + \lambda \Omega(f).
    \label{eq:cost}
\end{equation}

\begin{figure}[t]
    \centering
    \includegraphics[width=.8\linewidth]{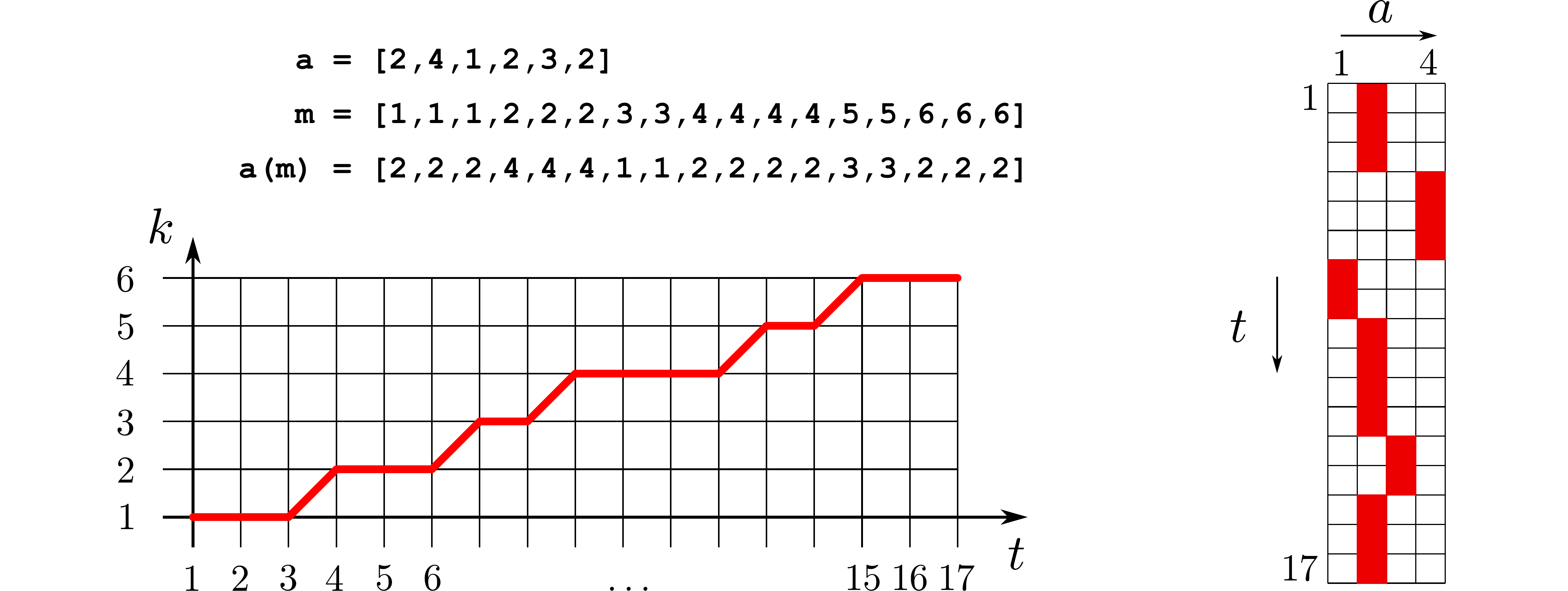}
    \caption{
        Illustration of the correspondence between temporal assignments (left) and associated valid assignment matrices that map action labels $a$ to time intervals $t$ (right). 
        \textbf{Left}: a valid assignment non-decreasing $m_t=k$. 
        \textbf{Right}: the corresponding assignment matrix $Z$. 
        One can build the assignment matrix \(Z\) given the assignment \(m\) and the annotation sequence \(a\) by putting a $1$ at index \((t,a(m_t))\) in $Z$ for every $t$. 
        One obtains \(m\) given $Z$ by iteratively constructing a sequence of integers of length \(T\) such that \(m_{t+1} = m_t\) if the $t$-th and $(t+1)$-th row of Z are identical, and \(m_{t+1} = m_t + 1\) otherwise. 
    }
    \label{fig:warping2}
\end{figure}


\subsection{Parameterization Using an Assignment Matrix}
\label{sec:indicators}

As will be shown in the following sections, it is convenient to reformulate our problem in terms of indicator variables.
The corresponding multi-class loss is \(\ell : \{0,1\}^A \times \mathbb{R}^A \to \mathbb{R}_+\), and the classifiers are functions \(f : \mathbb{R}^d \to \mathbb{R}^A\).
For a clip $n$, let us define the \emph{assignment matrix} \(Z^{n} \in \mathbb{R}^{T \times A}\) which is composed of entries \(z^{n}_{ta}\) such that \(z^{n}_{ta} = 1\) if the interval \(t\) of clip $n$ is assigned to class \(a\). 

Let \(Z^{n}_t\) denote the row vector of dimension \(A\) corresponding to the \(t\)-th row of \(Z^{n}\) .
The cost function \(E(m,f,n)\), defined in Eq.~\eqref{eq:E} can be rewritten as \(\frac{1}{T}\sum_{t=1}^{T}  \ell \left ( Z^n_t, f(x_n (t)) \right )\).

\textbf{Note:} To avoid cumbersome double summations, we suppose from now that we work with a single clip.
This allows us to drop the superscript notation, we replace \(Z^n\) by \(Z\) and skip the sum over clips.
We also replace the descriptor notation \(x_n(t)\) by \(x_t\) and the row extraction notation $Z^n_t$ by $Z_t$.
This is without loss of generality, and our method as described in the sequel handles multiple clips with some simple bookkeeping.

Because of temporal constraints, we want the assignment matrices \(Z\) to correspond to valid assignments \(m\). 
This amounts to imposing some constraints on~\(Z\). 
Let us therefore define \(\mathcal{Z}\), the set of all valid assignment matrices as:
\begin{equation}
    \mathcal{Z} = \left \{ Z  \in \{0,1\}^{T \times A} \ | \ \exists \ m \in \mathcal{M}, \ \text{s.t.}, \ \forall \ t, \ Z_{ta} = 1 \iff a(m_t) = a  \right \}.
\end{equation}
There is a bijection between the sets \(\mathcal{Z}\) and \(\mathcal{M}\). 
For each \(m\) in \(\mathcal{M}\) there exists a unique corresponding \(Z\) in \(\mathcal{Z}\) and \emph{vice versa}. 
Figure~\ref{fig:warping2} gives an intuitive illustration of this bijection.

The set $\mathcal{Z}$ is a subset of the set of stochastic matrices (positive matrices whose rows sum up to 1), formed by the matrices whose columns consist of exactly $K$ blocks of contiguous ones occurring in a predefined order ($K=6$ in Fig.~\ref{fig:warping2}).
There are as many elements in $\mathcal{Z}$ as ways of choosing $(K-1)$ transitions among $(T-1)$ possibilities, thus $\left | \mathcal{Z} \right | = \binom{T-1}{K-1}$, which can be extremely large in our setting (in our setting $T\approx100$ and $K\approx10$).
Furthermore, it is very difficult to describe explicitly the algebraic constraints on stochastic matrices that define $\mathcal{Z}$.
This point will prove important in Sec.~\ref{sec:optim} when we propose an optimization algorithm for learning our model.
Using these notations, Eq.~\eqref{eq:cost} is equivalent to:
\begin{equation}
    \min_{f \in \mathcal{F}, Z \in \mathcal{Z}} \ \frac{1}{T} \sum_{t=1}^T \ell \left ( Z_{t} , f (x_t) \right )  + \lambda \Omega(f).
    \label{eq:cost_in_Z}
\end{equation}


\subsection{Quadratic Cost Functions}
\label{sec:cost}
We now choose specific functions $\ell$ and $f$ that will lead to a quadratic cost function.
This choice leads, to a convex relaxation of our problem.
We use multi-class linear classifiers of the form \(f (x) = x^T W + b\), where \(W \in \mathbb{R}^{d \times A}\) and \(b \in \mathbb{R}^{1 \times A}\).
We choose the square loss function, regularized with the Frobenius norm of $W$, because in that case the optimal parameters \(W\) and \(b\) can be computed in closed form through matrix inversion.
Let \(X\) be the matrix in $\mathbb{R}^{T\times d}$ formed by the concatenation of all \(1 \times d\) matrices \(x_t\).
For this choice of loss and regularizer, our objective function can be rewritten using the matrices defined above as:
\begin{align}
    \frac{1}{T} \sum_{t=1}^T \ell(Z_t, f(x_t)) + \lambda \Omega(f)  = \frac{1}{T} \| Z - X W - b \|_F^2 + \frac{\lambda}{2} \|W\|_F^2.
    \label{eq:quad_cost_in_Z}
\end{align}
This is exactly a ridge regression cost.
Minimizing this cost with respect to \(W\) and \(b\) for fixed \(Z\) can be done in closed form~\cite{Bach07diffrac,Hastie2009elements}. 
Setting the partial derivatives with respect to \(W\) and \(b\) to zero and plugging the solution back yields the following equivalent problem:
\begin{equation}
    \min_{Z \in \mathcal{Z}} \ \text{Tr} \left (  Z Z^T B \right ), \ \text{where} \ B = \frac{1}{T} \Pi_T  ( I_T - X \left ( X^T \Pi_T X + T \lambda I_d \right )^{-1} X^T  ) \Pi_T,
    \label{eq:convex_in_Z}
\end{equation}
and the matrix \(\Pi_p\) is the \(p \times p\) centering matrix \(I_{p} - \frac{1}{p} \mathbf{1}_{p} \mathbf{1}_{p}^T\).
This corresponds to implicitly learning the classifier while finding the optimal Z by solving a quadratic optimisation problem in \(Z\).
The implicit classifier parameters \(W\) and \(b\) are shared among all video clips and can be recovered in closed-form as:
\begin{equation}
    W = (X^T \Pi_d X + \lambda  I)^{-1} X^T \Pi_T Z^*D^{1/2}, \qquad
    b =  \frac{1}{T} \mathbf{1}^T (Z^* - X w)D^{1/2}.
    \label{eq:Wb}
\end{equation}


\section{Convex Relaxation and the Frank-Wolfe Algorithm}
\label{sec:optim}

In Sec.~\ref{sec:formulation}, we have seen that our model can be interpreted as the minimization of a convex quadratic function ($B$ is positive semidefinite) over a very large but discrete domain. 
As is usual for this type of hard combinatorial optimization problem, we replace the discrete set $\mathcal{Z}$ by its convex hull $\overline{\mathcal{Z}}$. 
This allows us to find a continuous solution of the relaxed problem using an appropriate and efficient algorithm for convex optimization.


\subsection{The Frank-Wolfe Algorithm}
\label{sec:frank-wolfe}

We want to carry out the minimization of a convex function over a complex polytope
\(\overline{\mathcal{Z}}\), defined as the convex hull of a large but finite set of integer points defined by the constraints associated with admissible assignments.
When it is possible to optimize a linear function over a constraint set of this kind, but other usual operations (like projections) are not tractable, a good way to optimize a convex objective function is to use the iterative Frank-Wolfe algorithm (a.k.a. conditional gradient method) \cite{Bertsekas99nonlinear,Frank1956}.
We show in Sec.~\ref{sec:dynamicprogramming} that we can  minimize linear functions over \(\overline{\mathcal{Z}}\), so this is an appropriate choice in our case.

\begin{figure}[t]
    \centering
    \includegraphics[width=0.7\linewidth]{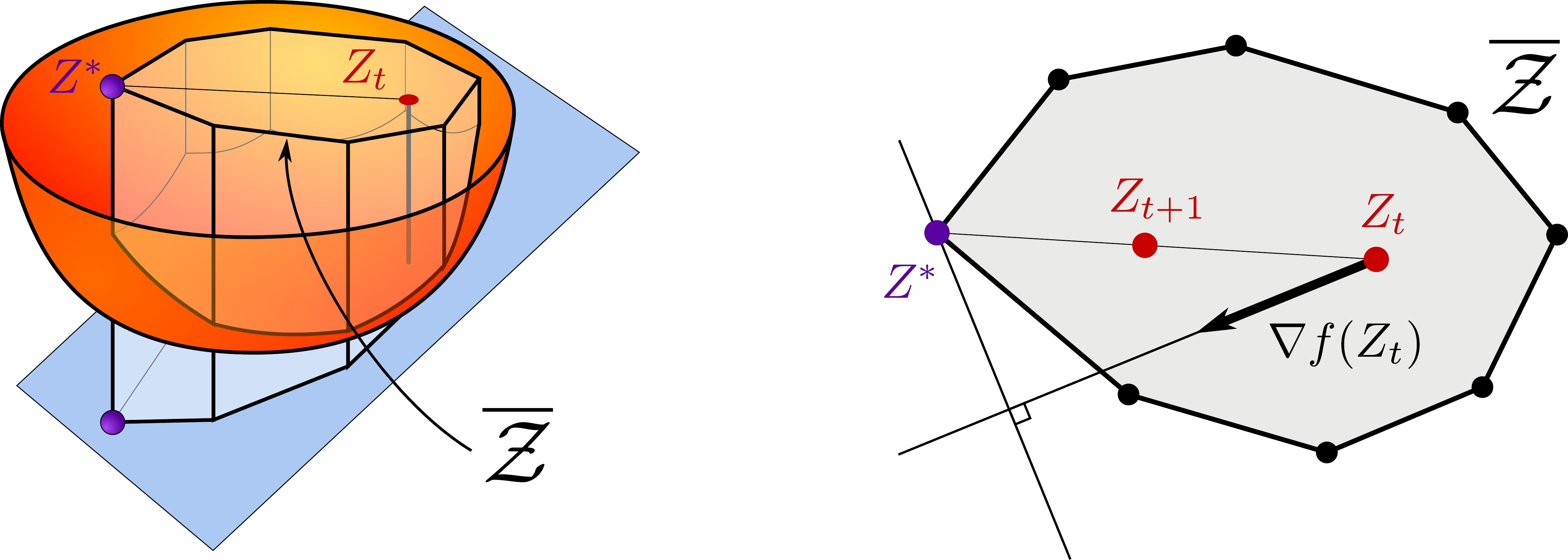}
    \caption{Illustration of a Frank-Wolfe step (see~\cite{Jaggi2013} for more details). \textbf{Left}: the domain $\overline{\mathcal{Z}}$ interest, objective function, and its linearization at current point. \textbf{Right}: top view of $\overline{\mathcal{Z}}$. Note that, in the algorithm, we actually minimize a linear function at each step. Adding a constant to it does not affect the solution of the minimization problem, it is equivalent to minimizing affine functions. That is why, we depicted an hyperplane that seems shifted from the origin.}
    \label{fig:frank-wolfe}
\end{figure}

The idea behind the Frank-Wolfe algorithm is rather simple. 
An affine approximation of the objective function is minimized yielding a point \(Z^*\) on the edge of $\overline{\mathcal{Z}}$.
Then a convex combination of $Z^*$ and the current point $Z$ is computed. This is repeated until convergence.
The interpolation parameter $\gamma$ can be chosen either by using the universal step size $\frac{2}{p+1}$, where $p$ is the iteration counter (see \cite{Jaggi2013} and references therein) or, in the case of quadratic functions, by solving a univariate quadratic equation.
In our implementation, we use the latter.
A good feature of the Frank-Wolfe algorithm is that it provides for free a duality gap (referred to as the linearization duality gap \cite{Jaggi2013}) that can be used as a certificate of sub-optimality and stopping criterion.
The procedure is described in the special case of our relaxed problem in Algorithm~\ref{alg:fw}.
Figure~\ref{fig:frank-wolfe} illustrates one step of the optimization.

\begin{algorithm}[b]
    $k \leftarrow 0$\\
    \While{$\rm{Tr}(\nabla_f(Z_k)(Z_k-Z^{*})) \geq \epsilon $}{
       	 Compute the current gradient in $Z,$ $\nabla_f(Z_k)=Z_k^{T}B.$ \\
   		 Solve $\min_{ Z \in \overline{\mathcal{Z}}} \text{Tr}(Z\nabla_f(Z_k))$ using dynamic programming.\\
   		 Compute the optimal Frank-Wolfe step size $\gamma.$ \\
   		 $Z_{k+1}=Z_{k}+\gamma(Z^{*}-Z_{k})$\\
   		     $k \leftarrow k+1$.

   		}
    	\caption{The Frank-Wolfe optimization procedure.}
        \label{alg:fw}
\end{algorithm}

\subsection{Linear Function Minimization over \(\overline{\mathcal{Z}}\)}
\label{sec:dynamicprogramming}

It is possible to minimize linear functions over the integral set \(\mathcal{Z}\). 
Simple arguments (see for instance Prop B.21 of \cite{Bertsekas99nonlinear}) show that the solution over \(\mathcal{Z}\) is also a solution over \(\overline{\mathcal{Z}}\).
We will therefore focus on the minimization problem on \(\mathcal{Z}\) and keep in mind that it also gives a solution over \(\overline{\mathcal{Z}}\) as required by the Frank-Wolfe algorithm.
Minimizing a linear function on \(\mathcal{Z}\) amounts to solving the problem:
$\label{eq:tracemin}
    \min_{Z \in \mathcal{Z}} \ \text{Tr} \left ( C^T Z \right )=\sum_{t=1}^T \sum_{a=1}^A \ Z_{ta} C_{ta},$
where \(C\) is a matrix in \(\mathbb{R}^{T\times A}\).
Using the equivalence between the assignment matrix ($Z$) and the plain assignment ($m$) representations (Fig.~\ref{fig:warping2}), this is equivalent to solving $\min_{m \in \mathcal{M}} \ \sum_{t=1}^T \sum_{a=1}^A \ \indicator{a(m_t) = a} C_{ta}$.
To better deal with the temporal structure of the assignment, let us denote by \(D \in \mathbb{R}^{T \times K}\) the matrix with entries $ D_{tk} = C_{t a(k)}$. 
The minimization problem then becomes \(\min_{m \in \mathcal{M}} \ \sum_{t=1}^T \ D_{t m_t}\), which can be solved using dynamic time warping.
Indeed, let us define for all \(t \in \{1,\dots,T\}\) and \(k \in \{1,\ldots,K\}\): $P_t^*(k) = \min_{m \in \mathcal{M}} \sum_{s=1}^t D_{s m_s}$.
We can think of $P_t^*(k)$ as the cost of the optimal path from $(1,1)$ to $(t,k)$ in the graph defined by admissible assignments, and we have the following dynamic programming recursion: $P_t^* (k) = D_{t k} + \min(P_{t-1}^* (k-1), P_{t-1}^* (k))$.

The optimal value $P_T^* (K)$ can be computed in $O(T K)$ using dynamic programming, by precomputing the matrix $D$, incrementally computing the corresponding $P_t^* (k)$ values, and maintaining at each node $(t,k)$ back pointers to the appropriate neighbors.


\subsection{Rounding}\label{sec:rounding}
At convergence, the Frank-Wolfe algorithm finds the (non-integer) global optimum \(Z^*\) of Eq.~\eqref{eq:convex_in_Z} over \(\overline{\mathcal{Z}}\).
Given \(Z^*\), we want to find an appropriate nearby point \(Z\) in \(\mathcal{Z}\).
The simplest geometric rounding scheme consists in finding the closest point of \(\mathcal{Z}\) according to the Frobenius distance : \(\min_{Z \in \mathcal{Z}} \| Z^* - Z \|_F^2\).
Expanding the norm yields: $\| Z^* - Z \|_F^2 = \text{Tr}  ( {Z^*}^T Z^*  )  + \text{Tr}  (  Z^T Z  ) - 2 \text{Tr}  ( {Z^*}^T Z  )$.

Since \(Z^*\) is fixed, its norm is a constant. Moreover, since \(Z\) is an element of \(\mathcal{Z}\), its squared norm is constant and equal to $T$. 
The rounding problem is therefore equivalent to: \(\min_{Z \in \mathcal{Z}} - 2 \text{Tr} ( {Z^*}^T Z )\), that is to the minimization of a linear function over $\overline{\mathcal{Z}}$.
This can be done, as in Sec.~\ref{sec:dynamicprogramming}, using dynamic programming.


\section{Practical Concerns}
\label{sec:extensions}
In this section, we detail some refinements of our model.
First we show how to tackle a semi-supervised setting where some time-stamped annotations are available. 
Secondly, we discuss how to avoid the trivial solutions, a common issue in discriminative clustering methods \cite{Joulin10discriminative,Bach07diffrac,Guo2007convex}.


\subsection{Semi-supervised Setting}
\label{sec:semi-supervised}
Let us suppose that we are given some fully annotated clips (in the sense that they are labeled with time-stamped annotations), corresponding to a total of \(L\) time intervals. 
For every interval \(l\) we have a descriptor \(X_l\) in \(\mathbb{R}^d\) and a class label \(a_l\) in \(\mathcal{A}\). 
We can incorporate this data by modifying the optimization problem as follows:
\begin{equation} 
    \min_{f \in \mathcal{F}} \ \left [ \min_{m \in \mathcal{M}} \ E(m,f,n) \right ] + \frac{1}{L} \sum_{l=1}^L \ell(a_l, f(X_l)) + \lambda \Omega(f).
\end{equation}

This supervised model does not change the optimization procedure, which remains valid.


\subsection{Minimum size constraints}

There are two inherent problems with discriminative clustering
First, the constant assignment matrix is typically a trivial optimum. 
As explained in \cite{Guo2007convex} this occurs when the optimization domain is symmetric over permutations of the labels of the assignment matrices. 
Due to our temporal constraints, the set $\mathcal{Z}$ is not symmetric and thus we are not subject to this effect.

The second difficulty is linked to the use of the centering matrix $\Pi_T$ in the expression of the quadratic cost matrix \(B\). 
Indeed, we notice that the constant vector of length \(T\) is an eigen vector of \(\Pi_T\). 
Therefore, the column-wise constant matrices are trivial solutions to our problem.
These piecewise-constant solutions are not admissible for our problem due to the temporal constraints.
In practice however, we have have observed that the algorithm returned an assignment with almost all points being affected to the background label $\varnothing$.
We consider two ways to get rid of the trivial solutions.

\subsection{Linear penalty.}
To avoid solutions with dominant classes we add constraints over the fraction of clip intervals affected to each class.
Ideally, we would like to incorporate a hard constraint over the proportions of each class as in \cite{Joulin10discriminative}, that is, to add to the problem formulated in Eq.~\eqref{eq:convex_in_Z}, a constraint of the type:
\begin{equation}
    \forall a \in \{1, \dots, A\} \ , \ n^a_\text{min} \leq \text{Tr} \left ( Z^T U_a \right ) \leq n^a_\text{max},
\end{equation} 
where $U_a \in \mathbb{R}^{n\times A}$ is the indicator matrix with 0 everywhere except on the $a$-th column which is 1.
This constraint would make all operations described in Sec.~\ref{sec:dynamicprogramming} intractable: indeed, dynamic programming cannot be modified so that it respects a constraint of minimal and maximal proportions.

Instead, a simple method for avoiding trivial solutions is to add to the objective function a Lagrangian multiplier corresponding to the desired hard constraints, which we will set by validation. 
We therefore incorporate a linear penalty (in $Z$) in our objective function.
The multiplier corresponding to this new term is defined as a vector $\kappa \in \mathbb{R}^{K}$. 
The final objective function then becomes:
\begin{equation}\label{eq:objpenalized}
    \min_{Z \in \overline{\mathcal{Z}}} \text{Tr} \left (  Z Z^T A \right ) +  \text{Tr} \left ( \kappa \mathbf{1}^T Z \right ).
\end{equation}
Note that, with this simple modification, we can still use Alg.~\ref{alg:fw}.

\subsection{Balancing the loss.}
Our constraint set is heavily unbalanced towards the \(\varnothing\) class.
A common way to deal with unbalanced datasets, is to weight the different classes appropriately:
Instead of considering in Eq.~\eqref{eq:quad_cost_in_Z} the standard least square regression problem, we propose to associate different weights to different labels.
If we denote by $D \in \mathbb{R}^{A \times A}$ the diagonal matrix containing the weights of each class, the square loss of Eq.~\eqref{eq:quad_cost_in_Z} becomes $\| (Z - X W - b)D \|_F^2 + \frac{\lambda}{2} \|WD\|_F^2.$
The actual values of $D$ are obtained by validation.
Note that this approach differs from so called re-weighted least squares (see for instance \cite{Hastie2009elements}), since here we weight labels and not instances.
Following  \cite{Bach07diffrac}, a simple computation shows that the matrix $B$ remains unchanged.
Thus, our algorithm is unchanged except in the computation of the Frank Wolfe gradient.


\section{Dataset and Features}
\label{sec:dataset}

\noindent\textbf{Dataset.}
Our input data consists of challenging video clips annotated with sequences of actions.
One possible source for such data is movies with their associated scripts~\cite{Bojanowski13finding,Duchenne09a,Laptev08a,Sivic09a}.
The annotations provided by this kind of data are noisy and do not provide ground-truth time-stamps for evaluation.
To address this issue, we have constructed a new action dataset, containing clips annotated by sequences of actions.
We have taken the 69 movies from which the clips of the Hollywood2 dataset were extracted~\cite{Laptev08a}, and manually added full time-stamped annotation for 16 classes (12 of these classes are already present in Hollywood2).
To build clips that form our input data, we search in the annotations for action chains containing at least two elements.
To do so, we pad the temporal action annotations by 250 frames and search for overlapping intervals.
A chain of such overlapping annotations forms one video clip with associated action sequence in our dataset.
In the end we obtain 937 clips, with number of actions ranging from 2 to 11. 
We subdivide each clip into temporal intervals of length 10 frames. 
Clips contain on average 84 intervals, the shortest containing 11, the longest 289.

\noindent\textbf{Feature representation.}
We have to define a feature vector for every interval of a clip.
We build a bag-of-words vector \(x_t\) per interval \(t\).
Recall that intervals are of length 10 frames. 
To aggregate enough features, we decided to pool features from the 30-frame-long window centered on the interval.
We compute video descriptors following~\cite{Wang13action}.
We generate vocabularies of size 2000 for HOF features. 
We restricted ourselves to one channel to improve the running time, while being aware that by doing so we sacrifice some performance.
In our informal experiments, we also tried the MBH channels yielding very close performance.
We use the Hellinger kernel to obtain the explicit feature map by square-rooting the \(l_1\) normalized histograms. 
Now every data point is associated with a vector \(x_t\) in \(\mathbb{R}^{2000}\).


\section{Action Labeling Experiments}
\label{sec:results}
\begin{figure}[t]
    \centering
    \includegraphics[width=0.4\linewidth]{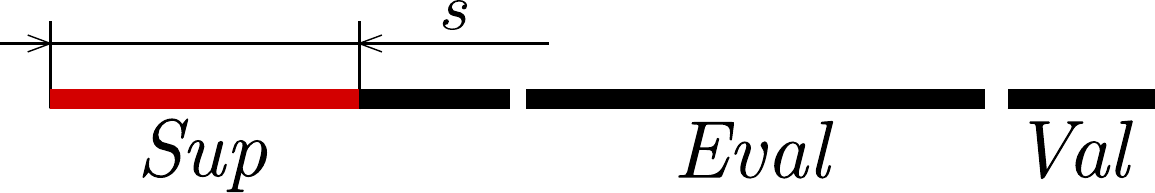}
    \caption{Splitting of the data described in Sec.~\ref{sec:spliting}.}
    \label{fig:data-split}
\end{figure}

\noindent\textbf{Experimental Setup.}
\label{sec:spliting}
To carry out the action labeling experiments, we split 90\% of the dataset into three parts (Fig.~\ref{fig:data-split}) that we denote \emph{Sup} (for supervised), \emph{Eval} (for evaluation) and \emph{Val} (for validation).
\emph{Sup} is the part of data that has time-stamped annotations, and it is used only in the semi-supervised setting described in Sec.~\ref{sec:semi-supervised}. 
\emph{Val} is the set of examples on which we automatically adjust the hyper-parameters for our method (\(\lambda, \kappa, D\)). 
In practice we fix the \emph{Val} set to contain 5\% of the dataset. 
This set is provided with fully time-stamped annotations, but these are not used during the cost optimization. 
None of the reported results are computed on this set. 
We evaluate the quality of the assignment on the \emph{Eval} set.
Note that we carry out the Frank-Wolfe optimization on the union of all three sets.
The annotations from the \emph{Sup} set are used to constrain \(Z\) in the semi-supervised setup while those from the \emph{Val} set are only used for choosing our hyper parameters. 
The supervisory information used over the rest of the data are the ordered annotations without time stamps.
Please also keep in mind that there are no ``training'' and ``testing'' phases \emph{per se} in this primary assignment task.
All our experiments are conducted over five random splits of the data. 
This allows us to present results with error bars.

\noindent\textbf{Performance Measure.}
Several measures may be used to evaluate the performance of discriminative clustering algorithms. 
Some authors propose to use the output classifier to perform a classification task~\cite{Duchenne09a,Xu2004maximum} or use the output partition of the data as a solution of the segmentation task~\cite{Joulin10discriminative}. 
Yet another way to evaluate is to use a loss between partitions~\cite{Hubert1985comparing} as in~\cite{Bach07diffrac}.
Note that because of temporal constraints, for every clip we have a set of corresponding (prediction, ground-truth) pairs.
We have thus chosen to measure the assignment quality for every ground-truth action interval \(I^\text{*}\) and prediction \(I\) as $|I \cap I^{*}|/|I|$.
This measure is similar to the standard Jaccard measure used for comparing ensembles \cite{Jaccard1912distribution}.
Therefore, with a slight abuse of notation, we refer to this measure as the Jaccard measure.
This performance measure is well suited for our problem since it respects the following properties: 
\textbf{(1)} it is high if the action predicted is included in the ground-truth annotation, 
\textbf{(2)} it is low if the prediction is bigger than the annotation, 
\textbf{(3)} it is lowest if the prediction is out of the annotation, 
\textbf{(4)} it does not take into account the prediction of the background class.
The score is averaged across all ground-truth intervals.
The perfect score of 1 is achieved when all actions are aligned to the correct annotations, but accurate temporal segmentation is not required as long as the predicted labels are within the ground truth interval.

\noindent\textbf{Baselines.}
We compare our method to the three following baselines. 
All these are trained using the same features as the ones used for our method.
For all baselines, we round the obtained solution \(Z\) using the scheme described in Sec.~\ref{sec:rounding}. 

\noindent \emph{Normalized Cuts (NCUT).}
We compare our method to normalized cuts (or spectral clustering)~\cite{Shi97normalized}. 
Let us define \(B\) as the symmetric Laplacian of the matrix \(E\): \(B = I - D^{-\frac{1}{2}} E D^{-\frac{1}{2}}\), where \(D = \text{Diag}\left ( E \mathbf{1} \right )\).
\(E\) measures both the proximity and appearance similarity of intervals. 
For all \((i,j)\) in \(\{1, \dots, T\}^2\), we compute: \(E_{ij} = e^{- \alpha |i-j| - \beta d_{\chi^{2}}(X_i,X_j)} \ \indicator{|i-j| < d_{\text{min}}} \), where \(d_{\chi^2}\) is the Chi-squared distance.
More precisely, we minimize over all cuts \(Z\) the cost $g (Z) = \text{Tr} \left ( Z Z^T B \right )$.
\(g\) is convex ($B$ is positive semidefinite) and we can use the Frank-Wolfe optimization scheme developed for our model.
Intuitively, this baseline is searching for a partition of the video such that time intervals falling into the same segments have close-by features according to the Chi-squared distance.

\noindent \emph{Bojanowski et al.~\cite{Bojanowski13finding}.}
We also consider our own implementation of the weakly-supervised approach proposed in~\cite{Bojanowski13finding}.
We replace our ordering constraints by the corresponding ``at least one'' constraints.
When an action is mentioned in the sequence, we require it appears at least once in the clip.
This corresponds to a set of linear constraints on \(Z\).
We adapt this technique in order to work on our dataset.
Indeed, the available implementation requires storing a square matrix of the size of the problem.
Instead, we choose to minimize the convex objective of~\cite{Bojanowski13finding} using the Frank-Wolfe algorithm which is more scalable.

\begin{figure}[t]
    \centering
    \includegraphics[width=0.33\linewidth]{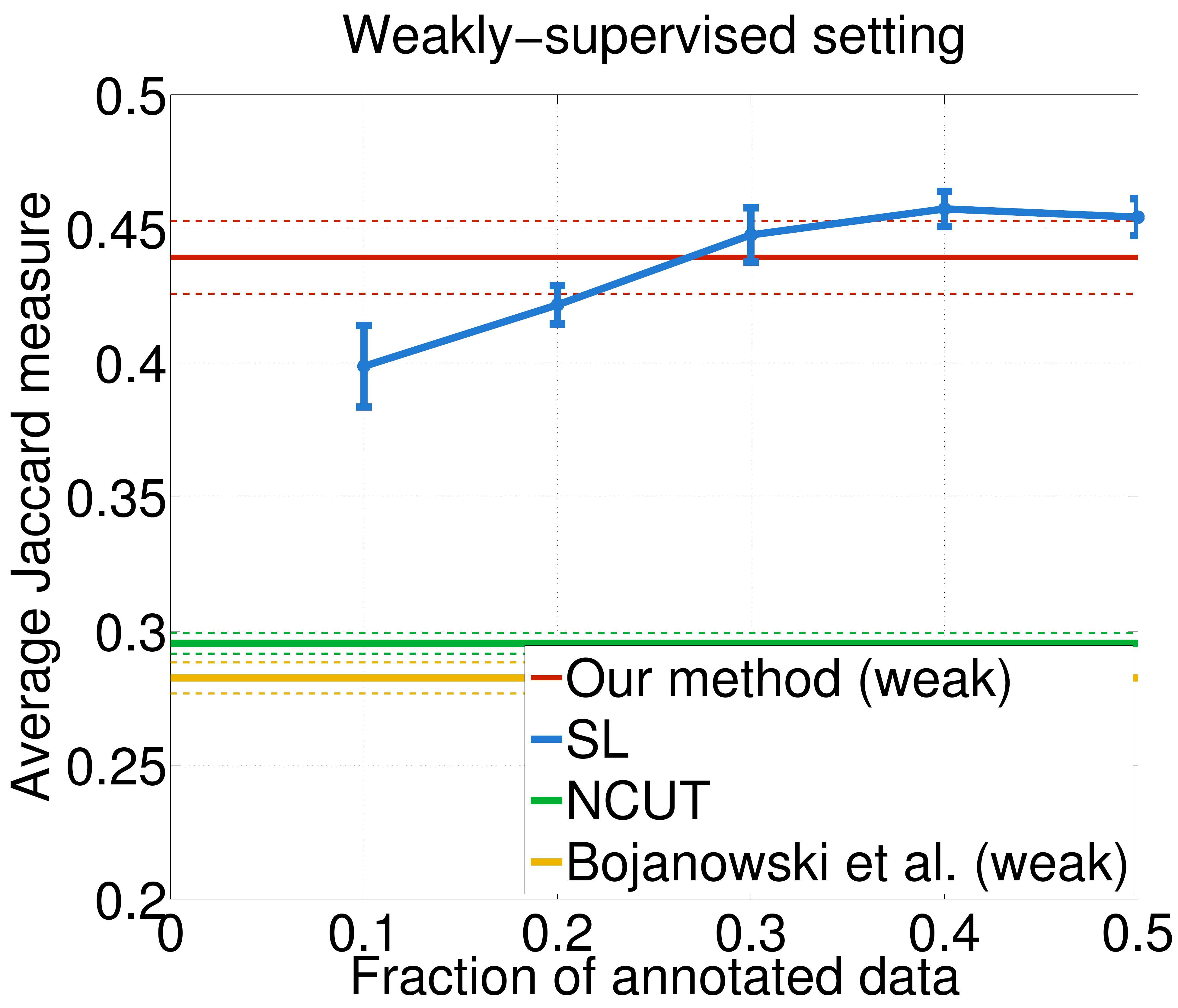}
    \includegraphics[width=0.33\linewidth]{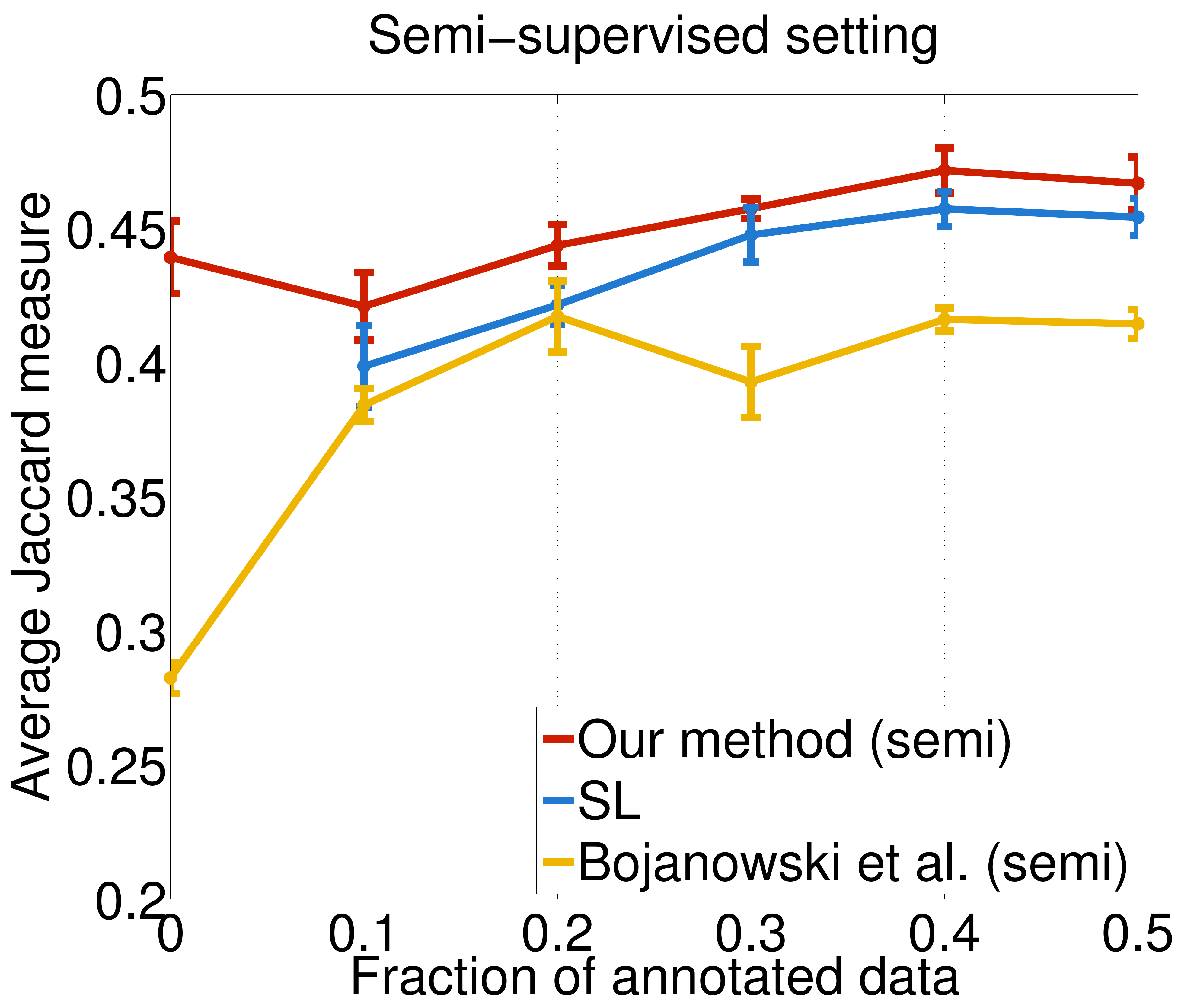}
    \caption{Alignment evaluation for all considered models. \textbf{Left}: weakly-supervised methods. This graph is shown for various fractions of fully supervised data only to compare to the SL baseline. Weak methods do not make use of this supervision. \textbf{Right}: semi-supervised methods. \textbf{See supplementary material for qualitative results.}}
    \label{fig:jac_s}
\end{figure}

\noindent \emph{Supervised Square Loss (SL).}
For completeness, we also compare our method to a fully supervised approach. 
We train a classifier using the square loss over the annotated \emph{Sup} set and score all time intervals in \emph{Eval}.  
We use the square loss since it is used in our method and all other baselines.

\begin{figure}[b]
    \centering
    \includegraphics[width=.8\linewidth]{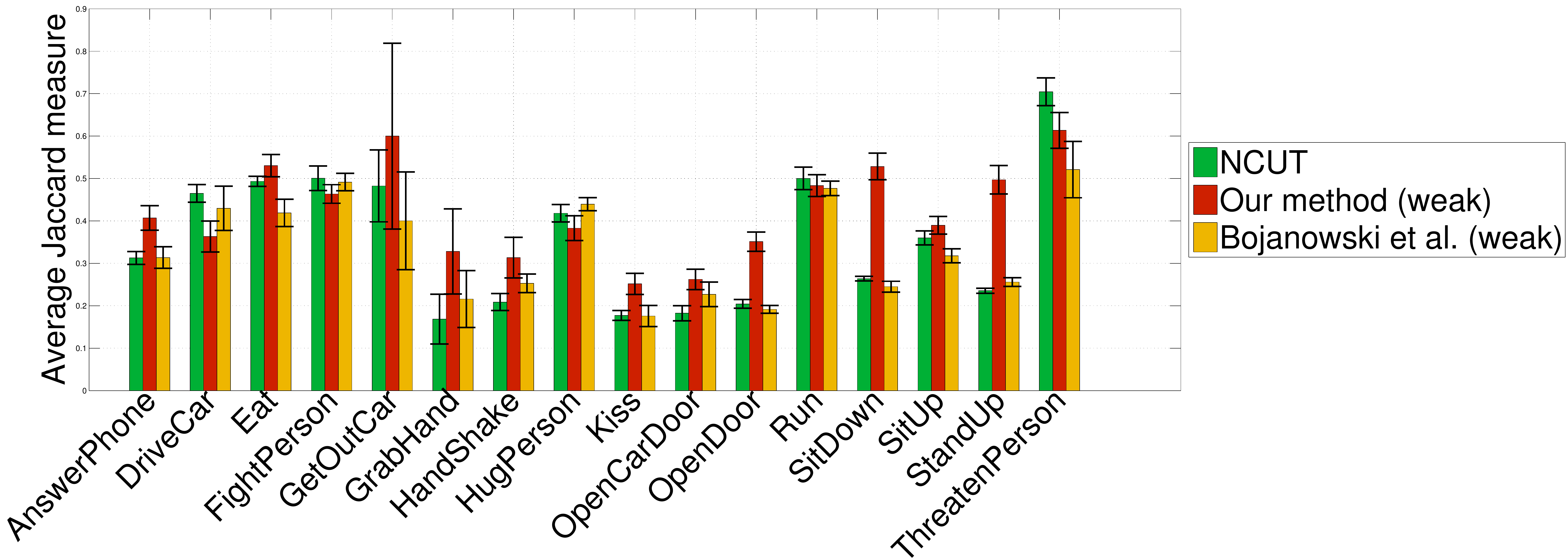}
    \caption{Alignment performance for various weakly-supervised methods for all classes.} 
    \label{fig:jac_barplot}
\end{figure}

\noindent\textbf{Weakly Supervised Setup.}
In this setup, all baselines except (SL) have only access to weak supervision in the form of ordering constraints.
Figure~\ref{fig:jac_s} (left) illustrates the quality of the predicted asignmentss and compares our method to baselines.
Our method performs better than all other weakly-supervised methods.
Both the Bojanowski et al. and NCUT baselines have low scores in the weakly-supervised setting. 
This shows the advantage of exploiting temporal constraints as weak supervisory signal.
The fully supervised baseline (blue) eventually recovers a better alignment than our method as the fraction of fully annotated data increases.
This occurs (when the red line crosses the blue line) at the 25\% mark, as the supervised data makes up for the lack of ordering constraints. 
Fully time-stamped annotated data are expensive to produce whereas movies scripts are often easy to get. 
It appears thus that manually annotated videos are not always necessary since good performance is reached simply by using weak supervision.
Figure~\ref{fig:jac_barplot} shows the results for all weakly-supervised methods for all classes.
We notice that we outperform the baselines on the most frequent classes (such as ``Open Door'', ``Sit Down'' and ``Stand Up'').

\noindent\textbf{Semi-supervised Setup.}
Figure \ref{fig:jac_s} (right) illustrates the performance of our model when some supervised data is available. 
The fraction of the supervised data is given on the x-axis.
First, note that our semi-supervised method (red) is always and consistently (Cf error bars) above the square loss baseline (blue). 
Of course, during the optimization, our method has access to weak annotations over the whole dataset, and to full annotations on the \emph{Sup} set whereas the SL baseline has access only to the latter. 
This demonstrates the benefits of exploiting temporal constraints during learning.
The semi-supervised Bojanowski et al. baseline (orange) has low performance, but it improves with the amount of full supervision provided. 


\section{Classification Experiments}\label{sec:expClassif}
The experiments in the previous section evaluate the quality of the recovered assignment matrix \(Z\).
Here we evaluate instead the quality of the recovered classifiers on a held-out test set of data for an action classification task.
We recover these classifiers as explained later in this section.
We can treat them as \(K\) independent, one-versus-rest classifiers and use them to score the samples from the test set.
We evaluate this performance by computing per-class precision and recall and report the corresponding average precision for each class.

\noindent\textbf{Experimental setup.}
The models are trained following the procedure described in the previous section.
To test the performance of our classifiers, we use the held out set of clips.
This set is made of 10\% of the clips from the original data.
The clips from this set are identical in nature to the ones used to train the models.
We also perform multiple random splits to report results with error bars.

\noindent\textbf{Recovering the classifiers.}
\label{sec:classifiers}
One of the nice features of our method is that we can estimate the implicit classifiers corresponding to our solution~\(Z^*\). 
We do so using the expression from Eq.~\ref{eq:Wb}.

\noindent\textbf{Baselines.}
We compare the classifiers obtained by our method to those obtained by the Bojanowski et al. baseline~\cite{Bojanowski13finding}.
We also compare them to the classifiers learned using the (SL) baseline.

\begin{figure}[b]
    \centering
    \includegraphics[width=0.33\linewidth]{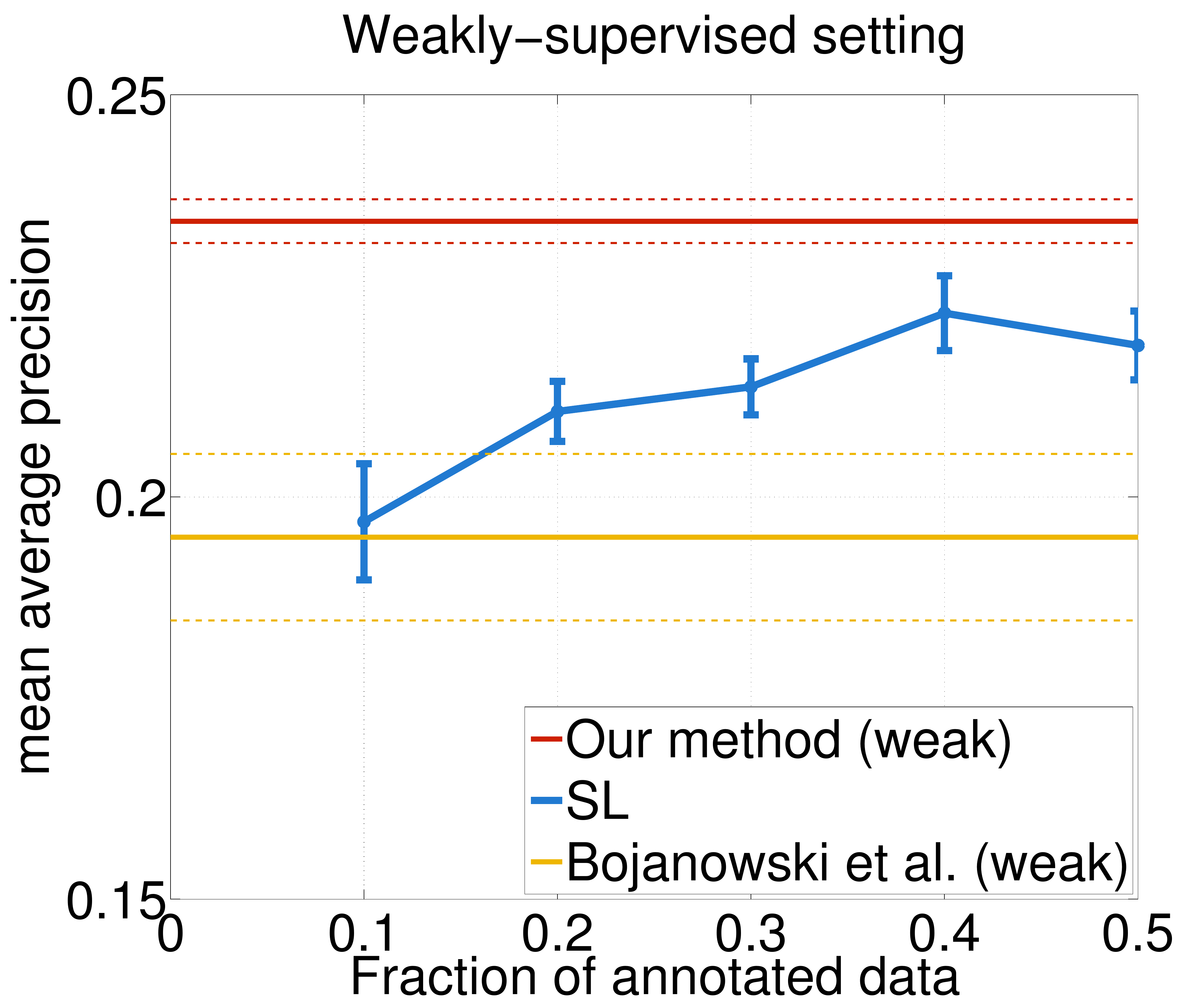}
    \includegraphics[width=0.33\linewidth]{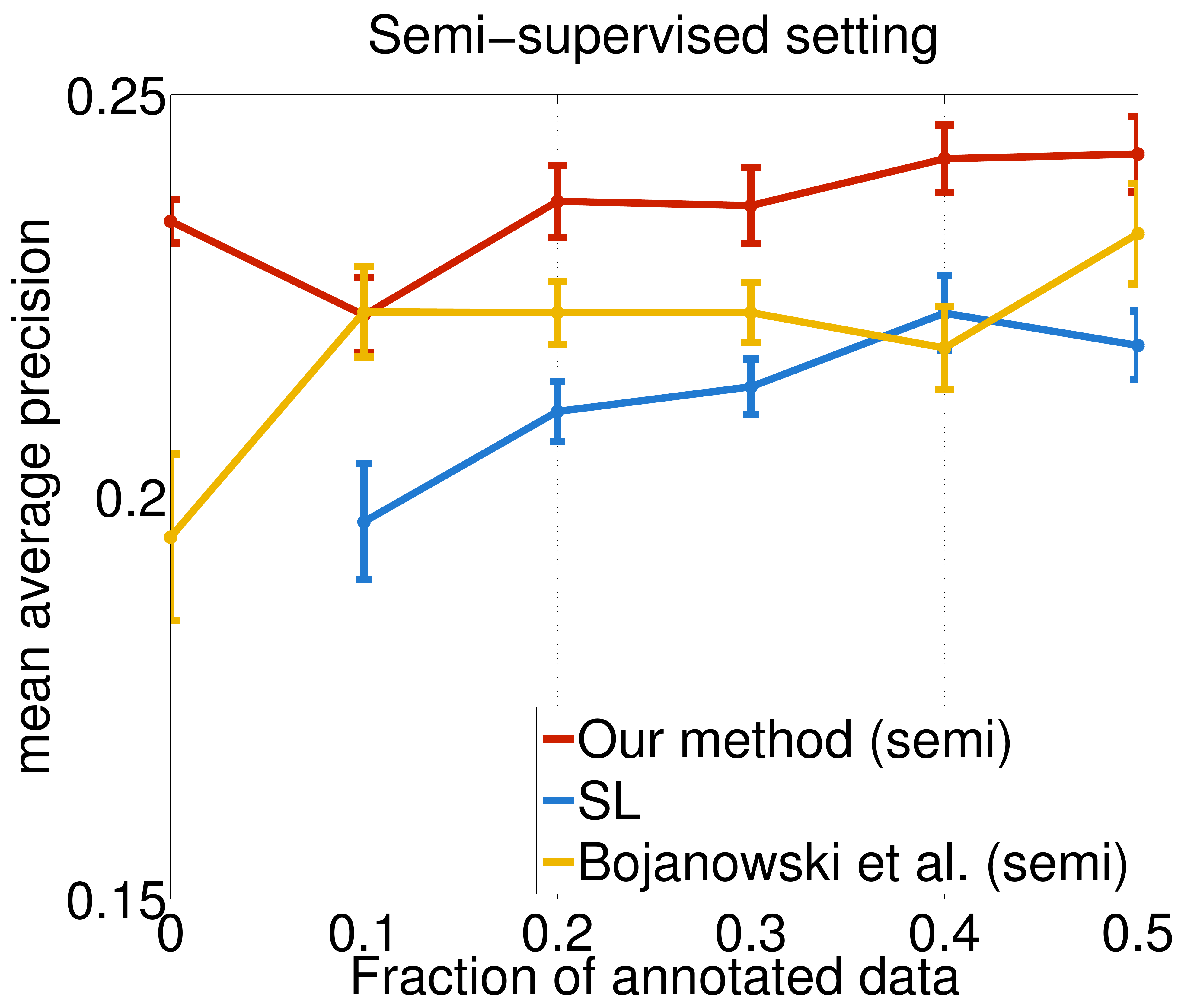}
    \caption{Classification performance for various models. \textbf{Left}: weakly-supervised methods. \textbf{Right}: semi-supervised methods. \textbf{Qualitative results in supp. material.}}
    \label{fig:classif_s}
\end{figure}

\noindent\textbf{Weakly Supervised Setup.}
Classification results are presented in Fig.~\ref{fig:classif_s} (left).
We observe a behavior similar to the action labeling experiment. 
But the supervised classifier (SL) trained on the \emph{Sup} set using the square loss (blue) always performs worse than our model (red).
This can be explained by the fact that the proposed model makes use of mode data.
Even though our model has only access to weak annotation, it can prove sufficient to train good classifiers.
The weakly-supervised method from Bojanowski et al. (orange) is performing worst, exactly as in the previous task.
This can be explained by the fact that this method does not have access to full supervision or ordering constraints.

\noindent\textbf{Semi-supervised Setup.}
In the semi-supervised setting (Fig.~\ref{fig:classif_s} (left)), our method (red) performs better than the supervised SL baseline (blue).
The action model we recover is consistently better than the one obtained using only fully supervised data.
Thus, our method is able to perform well semi-supervised learning.
The Bojanowski et al. baseline (orange) improves when the fraction of annotated examples increases.
Nonetheless, we see that making use of ordering constraints as used by our method signigicantly improves over simple linear inequalities (``at least one'' constraints as formulated in~\cite{Bojanowski13finding}).

\ \\

\noindent\textbf{Acknowledgements.}
\small{
This work was supported by the European integrated project AXES, the MSR-INRIA laboratory, EIT-ICT labs, a Google Research Award, a PhD fellowship from the EADS Foundation, the Institut Universitaire de France and ERC grants ALLEGRO, VideoWorld, Activia and Sierra. 
}

\bibliographystyle{splncs03}
\bibliography{main}
\end{document}